\documentclass[times, review, 10pt]{elsarticle}
\usepackage{amssymb}
\usepackage{amsmath}
\usepackage{bm}            
\usepackage{booktabs}      
\usepackage{graphicx}     
\usepackage{subcaption}
\usepackage{multirow}
\usepackage[ruled,vlined]{algorithm2e}
\usepackage{xcolor}
\usepackage{url}
\usepackage{float}

\journal{Neurocomputing}

\begin{document}

\begin{frontmatter}

\title{EdMCGS: Event-Driven Markov Chain Gaussian Splatting for Extreme-Low-Frame-Rate Dynamic Scene Reconstruction}

\author[mainaff]{Yuzhong Wang}
\author[mainaff]{Wenmin Wang\corref{cor1}}
\ead{wmwang@must.edu.mo}
\cortext[cor1]{Corresponding author}
\author[mainaff]{Xinxing Yu}

\affiliation[mainaff]{organization={Macau University of Science and Technology},
            city={Macau},
            country={China}}

\begin{abstract}
    We present EdMCGS (Event-driven Markov chain Gaussian Splatting), an end-to-end method for reconstructing dynamic 3D scenes from extreme-low-frame-rate RGB together with an event stream, which can then be rendered at any intermediate timestamp. Methods relying solely on RGB images generate numerous artifacts due to the lack of evidence from between consecutive frames. To supply this missing evidence, we model the scene motion as an event-driven Markov chain, in which the sparse RGB frames anchor the state at their own timestamps while the events recorded within an interval drive the transition across it. Since the transition reads the events of the current interval, it remains active at inference and produces the in-between motion of the 3D Gaussians directly from the events rather than by interpolation, which sets our method apart from prior work that uses events only as training-time supervision. The state is carried by a compact set of control points, each driven by the events sampled in the neighborhood of its own image projection, and a temporal local isometry term keeps the propagated motion locally rigid. Experiments on synthetic and real-world scenes show that EdMCGS outperforms both RGB-based and event-based baselines, while rendering in real time with far fewer Gaussians than the strongest event-based baseline. We release our source code and a new dataset at https://github.com/joseclipse/EdMCGS.
\end{abstract}




\end{frontmatter}

\section{Introduction}
\label{sec:intro}

High-quality 3D scene reconstruction and novel view synthesis have far-reaching applications across a wide range of domains, including autonomous driving, embodied intelligence and virtual reality. Neural Radiance Fields (NeRF) \cite{mildenhall2021nerf} demonstrated photorealistic rendering quality for static scenes, inspiring a substantial line of follow-up work that extended this framework to dynamic settings \cite{pumarola2021d, yan2023nerf, park2021hypernerf} and made the reconstruction of time-varying scenes possible. However, because NeRF-based methods rely on implicit functions to jointly encode scene geometry and appearance, they suffer from prohibitively slow rendering speeds and high memory consumption.

The recent advent of 3D Gaussian Splatting (3DGS) \cite{kerbl20233d} has enabled real-time, high-quality novel view synthesis by replacing the implicit volumetric representation with an explicit, point-based one, a collection of 3D Gaussian ellipsoids whose parameters are optimized end-to-end via differentiable rasterization. Building on this framework, 3DGS has quickly proven versatile across a broad range of tasks, from faster optimization \cite{hanson2025speedy, ren2025fastgs} to 3D scene segmentation \cite{shen2024flashsplat, ye2024gaussiangrouping}. Leveraging the expressiveness and efficiency of 3DGS, numerous approaches have further extended it to dynamic scenes, representing time-varying content as a set of deformable Gaussian primitives whose geometry and appearance evolve over time, as in Deformable 3DGS \cite{yang2024deformable} and 4D-GS \cite{wu_2024_CVPR}. Owing to fully differentiable, end-to-end optimization, 3DGS-based methods have rapidly surpassed NeRF-based methods on dynamic novel view synthesis. All of these methods nonetheless rely on densely sampled RGB input, where consecutive frames are close enough in time for the deformation field to interpolate the motion between them.

In many practical settings this assumption breaks down and the available RGB frames are sparse in time. Low frame rates arise from bandwidth and storage constraints, from long exposures in low-light capture, and from commodity hardware. Storage is an especially common pressure: full-frame-rate video is so bulky that it is often kept on a rolling buffer and overwritten after a short retention window, so that only recent footage can be retrieved. Keeping only a fraction of the RGB frames, for example one third to one sixth, together with the compact stream of an event camera is therefore an attractive way to retain reconstructable dynamic footage for far longer. Under such \emph{extreme-low-frame-rate} input, however, a model trained on RGB frames alone has no evidence about what happens between two consecutive frames; it tends to overfit the few observed timestamps and produces torn or ghosted geometry when asked to render the unobserved in-between motion. The core difficulty is not the representation but the lack of temporal evidence in the gaps between frames.

Event cameras offer exactly this missing evidence. As neuromorphic sensors, they asynchronously report per-pixel brightness changes at microsecond temporal resolution, with high dynamic range and low power consumption \cite{gallego2020event}. Rather than synchronous intensity frames at fixed intervals, an event camera produces a continuous stream that densely samples precisely the intervals where RGB frames are absent. This makes the event stream a natural complement to low-frame-rate RGB for dynamic reconstruction: the two modalities are deeply complementary, as RGB frames provide absolute appearance anchors at a coarse temporal resolution, while the event stream encodes fine-grained brightness changes that bridge the large temporal gaps between frames.

The promise of this complementarity has begun to be explored. A growing body of work reconstructs static scenes from events \cite{klenk2023enerf, yu2025evagaussians, kim2026e2egs}, and a handful of very recent methods target dynamic scenes \cite{he2025degs, xu2025eventboosted, nakabayashi2025ev4dgs, feng2025e4dgs}.
Yet across this nascent line three characteristics are common. First, the events shape the deformation only during training, where the deformation is parameterized as a function of time alone, so that at inference it is queried without events and the unobserved in-between motion is produced by temporal interpolation rather than driven by event evidence. Second, these methods tend to be heavy. DEGS \cite{he2025degs}, for instance, relies on a pretrained external event-flow network with per-scene low-rank fine-tuning, a geometry-aware event--Gaussian association, an explicit decomposition of scene and camera motion, and a separate pose network; other methods depend on dynamic-static or binary motion masks \cite{xu2025eventboosted, nakabayashi2025ev4dgs} or on multi-view event rigs \cite{feng2025e4dgs}. Finally, most of these methods and their datasets are not publicly available, and the one line whose code is released \cite{xu2025eventboosted} targets static-dominant captures, an assumption that we show empirically to leave much of the event evidence unexploited once the foreground motion dominates (Sec.~\ref{subsec:comparison}), so the community still lacks a reproducible baseline for the fully dynamic, extreme-low-frame-rate regime.

To address these limitations, we propose EdMCGS (Event-driven Markov chain Gaussian Splatting), a lightweight, end-to-end, and mask-free framework that reconstructs dynamic 3D scenes from sparse extreme-low-frame-rate RGB together with an event stream, and that can be rendered at any intermediate timestamp. Our central idea is to cast the motion as an event-driven Markov chain, in which the sparse RGB frames partition the timeline into intervals and anchor the state at their timestamps, while within each interval the event stream drives the transition from one state to the next. Because the transition is conditioned on the events, it remains active at inference, so that at an unobserved timestamp the in-between motion is produced directly from the events rather than guessed by interpolation, which is the key distinction from prior event-supervised methods. The state that the chain transitions is carried by a compact set of control points which drive the canonical Gaussians through linear blend skinning \cite{sumner2007embedded}, a low-rank motion representation that we adopt from SC-GS \cite{huang2024sc}. A temporal local isometry regularizer keeps the propagated motion locally rigid. Our main contributions are summarized as follows:

\begin{itemize}
  \item We propose EdMCGS, which models the motion of a dynamic scene as an \emph{event-driven Markov chain}, where the events recorded between two consecutive RGB frames drive the transition of the scene state from one frame to the next.

  \item We drive the chain by local event sampling, where every control point reads its own motion code from the accumulated event map at its image projection, so that the events fired by a moving part act on precisely the points covering that part instead of being collapsed into a single global descriptor.

  \item We introduce a temporal local isometry (TLI) regularizer that keeps the propagated motion locally rigid by preserving inter-point distances across nearby deformed timestamps, which stabilizes the rendering at unobserved in-between timestamps.

  \item We release our source code together with a new synthetic benchmark for event-based monocular dynamic scene reconstruction. On both synthetic and real-world scenes EdMCGS outperforms RGB-based and event-based methods at every frame rate we test, and it does so while training faster, rendering faster and using several times fewer Gaussians than the strongest event-based baseline.
\end{itemize}

\section{Related Work}
\label{sec:related}

\subsection{RGB-based Dynamic Scene Reconstruction}
\label{subsec:rw-rgb}

Novel view synthesis of dynamic scenes from RGB input has been studied intensively since Neural Radiance Fields (NeRF) \cite{mildenhall2021nerf}. One line of work extends NeRF to the temporal domain, either by conditioning an implicit field on a deformation that maps each observation back to a canonical space \cite{pumarola2021d, park2021hypernerf}, or by adopting explicit space-time structures such as planar or grid factorizations for efficiency \cite{Cao2023HexPlane, kplanes_2023, TiNeuVox}. Although these methods achieve high-quality interpolation, their implicit volumetric representation makes both training and rendering slow. 3D Gaussian Splatting (3DGS) \cite{kerbl20233d} instead represents the scene with an explicit set of anisotropic Gaussians that are rendered by differentiable rasterization, which enables real-time and high-fidelity synthesis. To extend 3DGS to dynamic scenes, some methods attach a time-conditioned deformation field to a canonical set of Gaussians \cite{yang2024deformable, wu_2024_CVPR, katsumata2024compact}, while others model the motion directly with 4D primitives or per-Gaussian trajectories \cite{luiten2024dynamic, kwak2025modec}. Among the deformation-based methods, several drive the dense Gaussians through a sparse set of learnable control points whose motion is regularized by an as-rigid-as-possible term \cite{huang2024sc, SP-GS}. More recently, MCGS \cite{wang2026mcgs} explored temporal motion propagation based on a Markov chain over control points, where the transition is predicted from a memory of past states with temporal attention and therefore still relies on densely sampled RGB observations. Our method keeps that sparse control point formulation, but addresses a different problem with a different transition mechanism. A related line of work tackles intra-frame motion blur in dynamic 3DGS \cite{wu2026deblur4dgs, lu2025bard, bui2026mobgs}, which is orthogonal to our setting because we address the inter-frame gaps left by sparse but sharp frames rather than blur within a single exposure.

\subsection{Event-based Static Scene Reconstruction}
\label{subsec:rw-event-static}

Event cameras report asynchronous, per-pixel brightness changes at microsecond resolution with high dynamic range \cite{gallego2020event}. Beyond low-level sensing, they have also been used for higher-level vision tasks such as object recognition \cite{LIU2025neuro}, motion feature extraction \cite{LIU2025An}, and knowledge transfer across event domains \cite{LIN2026ffevent}, and a large body of work has incorporated them into static 3D reconstruction. Within the NeRF framework, E-NeRF \cite{klenk2023enerf} couples the event generation model with volumetric rendering, and later work improves its robustness to event noise and varying contrast thresholds \cite{low2023robuste-nerf}, to inaccurate poses and uneven event density \cite{feng2025aenerf}, and to motion blur \cite{low2024deblure-nerf, qi2023e2nerf}. More recently, the explicit and efficient nature of 3DGS has been combined with events for static scenes, reconstructing from the event stream on its own or together with blurry or sparse RGB frames, as in recent event-based 3DGS methods \cite{wu2024evgs, yu2025evagaussians, huang2025inceventgs}. These methods show that the event stream is a powerful geometric cue, but because they assume a static scene they do not model the temporal deformation that is central to our problem.

\subsection{Event-based Dynamic Scene Reconstruction}
\label{subsec:rw-event-dynamic}

Reconstructing dynamic scenes with events is a nascent and rapidly moving direction, and it is here that our work is positioned. DE-NeRF \cite{qi2023denerf}, which first combined events with RGB for a dynamic radiance field, estimates a per-event color and jointly optimizes a deformable NeRF, although it inherits the slow rendering of NeRF and uses the events only as a supervision signal. DEGS \cite{he2025degs} brings event-guided deformation to 3DGS by extracting event-flow trajectories with a pretrained external flow estimator that is fine-tuned per scene through low-rank adaptation, by building a geometry-aware association between events and Gaussians from unprojected depth, and by decomposing object motion from camera ego-motion with a separate pose network. Although effective, this pipeline is heavy because it depends on several external components. Event-boosted Deformable 3DGS \cite{xu2025eventboosted} focuses instead on modeling the event contrast threshold and on a dynamic-static decomposition for faster rendering, which requires explicit dynamic-region masks. Ev4DGS \cite{nakabayashi2025ev4dgs} targets non-rigid reconstruction from a monocular event stream alone, using a low-rank deformation basis together with binary masks generated from the events, while E-4DGS \cite{feng2025e4dgs} addresses the multi-view setting in which the scene is captured by a $360^{\circ}$ rig of event cameras.

Across this line, the events serve only as training-time supervision, so at inference the in-between motion is interpolated rather than driven by evidence, and, apart from \cite{xu2025eventboosted}, neither the code nor the data is released. In contrast, EdMCGS reconstructs dynamic scenes from sparse RGB and a monocular event stream by casting the deformation as an event-driven Markov chain over control points whose transitions remain active at inference, and it is lightweight and mask-free, requiring no external optical-flow model, no per-scene fine-tuning, and no pose network. We release our code and dataset as a reproducible baseline for the extreme-low-frame-rate setting.

\section{Method}
\label{sec:method}

\begin{figure}[t]
  \centering
  \includegraphics[width=\columnwidth]{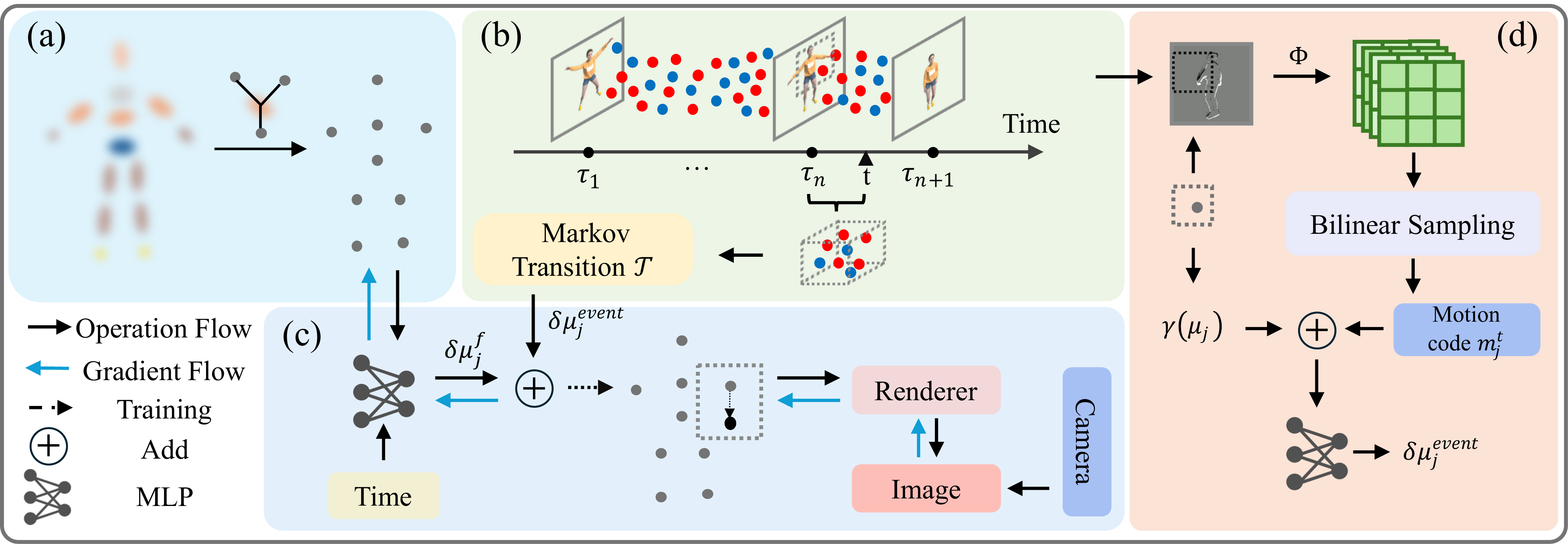}
  \caption{Overview of EdMCGS. (a)~We initialize the control points by sampling from the canonical space. (b)~We cast the deformation as an event-driven Markov chain, where the events accumulated in the interval $[\tau_n,t]$ between two RGB anchors drive a Markov transition of the control point state. (c)~At a query time $t$, the coarse RGB-driven deformation predicted by the MLP and the fine event-driven correction from the Markov transition are added to displace the control points, which in turn deform the Gaussians by linear-blend skinning before rasterization. (d)~Local event sampling: the events in $[\tau_n,t]$ are accumulated into a polarity map, encoded by a shallow CNN $\Phi$, and bilinearly sampled at each control point's projection to form its local motion code.}
  \label{fig:pipeline}
\end{figure}

We propose Event-driven Markov chain Gaussian Splatting (EdMCGS), shown in Fig.~\ref{fig:pipeline}, a new method for reconstructing a dynamic 3D scene from an extreme-low-frame-rate RGB sequence and an asynchronous event stream, so that the scene can be rendered at any intermediate timestamp not covered by an RGB frame. EdMCGS represents the scene as a canonical set of 3D Gaussians whose motion is carried by a sparse set of control points, and its central idea is to cast the deformation as an event-driven Markov chain in which the sparse RGB frames anchor the control points' state at their timestamps while the event stream drives the transition within each interval. Because the transition is driven by the events, it remains active at inference, so that the motion at the unobserved in-between timestamps is produced directly from the events rather than interpolated from the RGB frames.

In this section, we first briefly review the necessary background to establish the foundation for our approach (Sec.~\ref{subsec:prelim}). We then introduce our event-driven Markov chain (Sec.~\ref{subsec:markov}), followed by local event sampling (Sec.~\ref{subsec:local}), the temporal local isometry regularizer (Sec.~\ref{subsec:tli}) and the training objective (Sec.~\ref{subsec:loss}).

\subsection{Preliminaries}
\label{subsec:prelim}

\subsubsection{Deformable 3D Gaussian Splatting}
\label{subsubsec:prelim-3dgs}

3D Gaussian Splatting (3DGS) \cite{kerbl20233d} represents a scene by a set of $N$ anisotropic Gaussians $\mathcal{G}=\{G_i\}_{i=1}^{N}$. At time $t$, Gaussian $i$ is parameterized by a center $\mu_i^{t}\in\mathbb{R}^3$, a rotation quaternion $q_i^{t}$, a scale $s_i^{t}\in\mathbb{R}^3$, an opacity $\sigma_i$, and spherical-harmonic color coefficients $sh_i$, where the last two are time-invariant. The spatial response is
\begin{equation}
\begin{aligned}
  G_i(x)&=\exp\!\Big(-\tfrac{1}{2}(x-\mu_i^{t})^{\!\top}\Sigma_i^{-1}(x-\mu_i^{t})\Big),\\
  \Sigma_i&=R(q_i^{t})\,S(s_i^{t})\,S(s_i^{t})^{\!\top}R(q_i^{t})^{\!\top},
\end{aligned}
  \label{eq:gaussian}
\end{equation}
with $R(\cdot)$ the rotation matrix and $S(\cdot)$ the diagonal scale matrix. Images are formed by projecting the Gaussians to 2D image plane and $\alpha$-blending them front to back, which is fully differentiable and real-time.

To model a dynamic scene, Deformable 3DGS \cite{yang2024deformable} keeps $\mathcal{G}$ as a \emph{canonical} (time-zero) configuration and predicts, for a normalized time $t\in[0,1]$, a per-Gaussian deformation
\begin{equation}
  (\delta\mu_i,\delta q_i,\delta s_i)=f_\theta(\gamma(sg(\mu_i)),\gamma(t))
  \label{eq:d3dgs}
\end{equation}
that maps each canonical Gaussian to its state at $t$, where $sg(\cdot)$ indicates a stop-gradient operation and $\gamma(\cdot)$ is the standard sinusoidal positional encoding. This MLP must be evaluated for all $N\!\sim\!10^{5}$ Gaussians at every step, and, more importantly for our setting, under extreme-low-frame-rate input it overfits the few observed timestamps and tears apart at unobserved ones. This is the starting point of our method.

\subsubsection{Markov Chain}
\label{subsubsec:prelim-markov}

A discrete-time stochastic process $\{S_t \mid S_t \in S \;\text{and}\; t \in T\}$ is a first-order Markov chain if its future is conditionally independent of its past given the present,
\begin{equation}
  P(S_{t+1}=s_{t+1} \mid S_t=s_t, S_{t-1}=s_{t-1}, \ldots, S_1=s_1) = P(S_{t+1}=s_{t+1} \mid S_t=s_t),
  \label{eq:markovdef}
\end{equation}
so that the evolution of the process is fully described by its current state and a transition operator. In Sec.~\ref{subsec:markov} we instantiate the state as the configuration of control points and the transition as an event-driven operator, so that the motion within an interval is propagated from the preceding RGB frame by the events alone.

\subsubsection{Event Generation Model}
\label{subsubsec:prelim-egm}

An event camera reports asynchronous events $e=(x,y,t,p)$, with pixel location $u=(x,y)$, timestamp $t$, and polarity $p\in\{+1,-1\}$. An event is fired whenever the log-luminance $L(u,t)=\log\big(\mathcal{I}(u,t)+\epsilon\big)$ at a pixel changes by a contrast threshold $C$. Consequently, integrating the signed events over an interval $[t_a,t_b]$ predicts the log-luminance difference,
\begin{equation}
  L(u,t_b)-L(u,t_a)=C\!\!\sum_{e\in[t_a,t_b]}\!\! p_e,
  \label{eq:egm}
\end{equation}
which is the event generation model (EGM). It links a rendered brightness change to the accumulated events and is the basis of our event supervision.

\subsection{Event-Driven Markov Chain}
\label{subsec:markov}

We begin by sampling a sparse set of points from the canonical Gaussians by farthest point sampling, which we define as a set of control points $\mathcal{M}=\{M_j\}_{j=1}^{M}\subset\mathcal{G}$, inspired by the sparse control points in SC-GS \cite{huang2024sc}. These control points are connected to their nearby Gaussians by $K$-nearest-neighbor search and drive the deformation of those Gaussians through linear-blend skinning \cite{sumner2007embedded}, weighting each Gaussian by a Gaussian kernel of its distance to the control point. While control points are widely used \cite{huang2024sc, SP-GS, wang2026mcgs}, our contribution is not the control points themselves but how they are moved.

Let the extreme-low-frame-rate RGB frames be captured at the sparse timestamps $\tau_1<\tau_2<\dots<\tau_{N_{\mathrm{rgb}}}$, which partition $[0,1]$ into intervals. We take the state of the Markov chain at time $t$ to be the configuration of control points $S^{t}=\{\mu_j^{t}\}_{j=1}^{M}$, namely the deformed control point positions. For a query time $t$, let $\tau(t)=\max\{\tau_n:\tau_n\le t\}$ denote the immediately preceding RGB anchor. Since an RGB frame is observed at each $\tau_n$, the state at an anchor is well constrained by the photometric loss, whereas the in-between states must be inferred.

We model the evolution between consecutive anchors as a first-order Markov transition: the state at a query time $t$ depends on the past only through the preceding anchor state $S^{\tau(t)}$ and the events $E[\tau(t),t]$ observed in between, without any earlier history,
\begin{equation}
  S^{t}=\mathcal{T}\big(S^{\tau(t)},\,E[\tau(t),t]\big).
  \label{eq:transition}
\end{equation}
This instantiates the Markov property of Eq.~\eqref{eq:markovdef}, with the RGB frames acting as the anchors that pin the state and the events driving the transition operator $\mathcal{T}$ within each interval, so that the deformation remains active at inference.

We realize this transition independently for each control point. To the RGB prior $\delta\mu_j^{f,t}$, obtained by evaluating the deformation network of Eq.~\eqref{eq:d3dgs} at the $M$ control points rather than at all $N$ Gaussians so that the RGB-driven motion stays low-dimensional, we add an event-driven term, so that the total displacement of control point $j$ is
\begin{equation}
  \delta\mu_j^{t}=\delta\mu_j^{f,t}+g_\phi\big(\gamma(\mu_j),\,m_j^{t}\big),
  \label{eq:total}
\end{equation}
where $g_\phi$ is a small MLP, and $m_j^{t}$ is a motion code built from the events accumulated since the anchor, defined below. Since $\delta\mu_j^{f,t}$ is supervised only by the RGB frames, it is a smooth interpolator in $t$ that gives a sensible baseline near the observed timestamps but carries no evidence of the true motion in the gaps between sparse frames. The Markov transition is thus realized entirely by the event term $g_\phi$: at $t=\tau(t)$ the accumulation window is empty and the zero-initialized $g_\phi$ vanishes, so that the state reduces to the anchor $S^{\tau(t)}$, and as $t$ advances the accumulated events propagate it forward.

The canonical control points $\{\mu_j\}$ are fixed, so the anchor state $S^{\tau(t)}$ is a deterministic function of $\tau(t)$, which is in turn determined by $t$; conditioning on $(\mu_j,t)$ is therefore informationally equivalent to conditioning on $S^{\tau(t)}$, and the two parameterizations describe the same operator. What makes the process Markovian is not the parameterization of the prior but the accumulation window of the motion code, since $m_j^{t}$ reads only the events fired in $[\tau(t),t]$. The increment that carries the state from the anchor to $t$ therefore depends on the current interval alone and is conditionally independent of everything that occurred before $\tau(t)$, which is exactly the property required by Eq.~\eqref{eq:markovdef}. Each RGB frame resets the window, so errors accumulated inside one interval are not propagated across the anchors, a property that a transition conditioned on a memory of past states does not have.

\subsection{Local Event Sampling}
\label{subsec:local}

It remains to define the motion code $m_j^{t}$ that drives the transition, which we build from the interval events in three steps. First, the events fired in the anchored window $[\tau(t),\,t]$ are accumulated, by polarity, into a 2D map $I_E^{t}=\mathrm{Acc}\big(E[\tau(t),t]\big)\in\mathbb{R}^{H\times W}$ on the event camera image plane. Second, a shallow CNN $\Phi$ encodes this map into a feature map that preserves its spatial layout. Third, we project each control point onto the event camera and read its feature by bilinear sampling at that location,
\begin{equation}
  m_j^{t}=\mathrm{bilinear}\Big(\Phi\big(I_E^{t}\big),\;
  \pi_{\mathrm{ev}}\big(\mu_j+\delta\mu_j^{f,t}\big)\Big),
  \label{eq:motioncode}
\end{equation}
where $\pi_{\mathrm{ev}}$ is the projection onto the event camera and $\mu_j+\delta\mu_j^{f,t}$ is the prior-deformed control point.

Each control point samples the event feature in the neighborhood of its own image projection, rather than from a global summary of the event stream. Because the events triggered by a moving part fall near the projection of the control points that cover it, this local sampling lets the events of a region drive exactly the control points of that region, instead of being averaged together with camera and background events as in a global pooling. The neighborhood is defined implicitly by the receptive field of $\Phi$ and the bilinear sampling, without any explicit radius.

The branch $g_\phi$ is zero-initialized at its last layer and warmed in gradually, so optimization first establishes the RGB-anchored field before the events contribute, and sharing $g_\phi$ across control points together with the local sampling in Eq.~\eqref{eq:motioncode} keeps the transition spatially coherent, which is further encouraged by a KNN Laplacian smoothness term (Sec.~\ref{subsec:loss}).
\subsection{Temporal Local Isometry Regularization}
\label{subsec:tli}

To keep the propagated motion locally rigid without imposing a global rigid model, we regularize the deformation to be locally isometric in time. On the canonical control points we build a KNN graph with edge set $\mathcal{E}$. For an edge $(j,k)\in\mathcal{E}$, let 
\begin{equation}
  \ell_{jk}(t)=\lVert\mu_j^{t}-\mu_k^{t}\rVert
  \label{eq:ljk}
\end{equation}
be its deformed length at time $t$. At a sampled query $t$ we take the two nearby times $t_{1,2}=t\mp\tfrac{\Delta t}{2}$ and penalize the change of edge length between them,
\begin{equation}
  \mathcal{L}_{\mathrm{TLI}}=\frac{1}{|\mathcal{E}|}\sum_{(j,k)\in\mathcal{E}}
  \big(\ell_{jk}(t_1)-\ell_{jk}(t_2)\big)^{2}.
  \label{eq:tli}
\end{equation}
A locally rigid motion preserves inter-point distances and is therefore not penalized, while only stretching and shearing are. We stress that $\mathcal{L}_{\mathrm{TLI}}$ differs in substance from the ARAP regularizer of SC-GS, and not merely in name. First, it enforces local isometry, namely the preservation of inter-point distances, rather than fitting a per-point rotation by SVD and penalizing the residual, and it requires no rotation estimation at all. Second, it compares two deformed timestamps ($t_1$ and $t_2$) against each other, rather than comparing a deformed configuration against the static canonical one. The result is a rotation-free term that constrains how the geometry evolves between adjacent instants, which is exactly the property needed for stable in-between interpolation.

\subsection{Training Objective}
\label{subsec:loss}

\paragraph{Photometric loss}
At each RGB timestamp the rendering is supervised with the standard photometric loss, a combination of an $\ell_1$ term and a D-SSIM term.

\paragraph{Event loss}
We supervise the geometry and motion in the gaps with the EGM of Eq.~\eqref{eq:egm}. We render the scene at two dense sub-step times $t_a<t_b$, convert each rendering to log-luminance $\hat{L}=\log(\hat{\mathcal{I}}+\epsilon)$, and match the rendered difference to the accumulated events,
\begin{equation}
  \mathcal{L}_{\mathrm{event}}=
  \Big\lVert\,\big(\hat{L}(t_b)-\hat{L}(t_a)\big)
  - C\!\!\sum_{e\in[t_a,t_b]}\!\! p_e\,\Big\rVert_1 .
  \label{eq:event}
\end{equation}
Since a single brightness-change constraint cannot disentangle geometry from chromaticity, we \emph{detach the color gradient} when forming $\hat{L}$, so that $\mathcal{L}_{\mathrm{event}}$ updates only geometry and motion and does not contaminate the Gaussian colors, an issue we observed when the events were allowed to back-propagate into color.

\paragraph{Full objective}
Together with the two regularizers, the full objective is
\begin{equation}
  \mathcal{L}=(1-\lambda_d)\,\mathcal{L}_1
  +\lambda_d\,\mathcal{L}_{\mathrm{D\text{-}SSIM}}
  +\lambda_e\,\mathcal{L}_{\mathrm{event}}
  +\lambda_t\,\mathcal{L}_{\mathrm{TLI}}
  +\lambda_s\,\mathcal{L}_{\mathrm{smooth}},
  \label{eq:totalloss}
\end{equation}
where $\mathcal{L}_{\mathrm{smooth}}$ is the KNN Laplacian smoothness on the event-driven transition term $g_\phi$ that keeps neighboring control points coherent, and $\lambda_d,\lambda_e,\lambda_t,\lambda_s$ balance the terms.

\section{Experiments}
\label{sec:exp}

\subsection{Datasets}
\label{subsec:datasets}

\paragraph{Synthetic dataset}
We construct a synthetic event-based monocular dataset in Blender, consisting of four sequences, \emph{Jumpingjack}, \emph{Kick}, \emph{Mutant} and \emph{Lego}, which together cover articulated full-frame motion, localized limb motion and rigid mechanical motion. For each sequence we simulate a continuous camera trajectory around a deformable object and render a high-frame-rate RGB sequence with motion blur enabled at a shutter setting of $0.5$ ($180^\circ$ shutter angle), so that the exposure of fast motion is realistic. All sequences are rendered at $346 \times 260$ to match the resolution of a DAVIS346 event camera, and the rendered sequence is fed to ESIM \cite{rebecq2018esim} with a contrast threshold of $0.2$ to obtain the event stream. To emulate extreme-low-frame-rate capture we retain only the frames sampled at $10$ fps or $5$ fps for training, which amounts to roughly $20$ and $10$ input frames per sequence, and the discarded frames serve as ground truth at the unobserved intermediate timestamps.

\paragraph{Real-world dataset}
For real captures we use the real-world dataset released by E-D3DGS \cite{xu2025eventboosted}, which provides four dynamic scenes, \emph{Excavator}, \emph{Jeep}, \emph{Flowers} and \emph{Eagle}, recorded with a real event camera together with the RGB frames. We downsample each RGB stream to $5$ fps to match our extreme-low-frame-rate setting and hold out the remaining frames as ground truth at the intermediate timestamps.

\subsection{Experimental Setup}
\label{subsec:setup}

\paragraph{Baselines}
We compare EdMCGS against two RGB-based deformable baselines, Deformable 3DGS (D-3DGS) \cite{yang2024deformable} and SC-GS \cite{huang2024sc}. Since RGB-based methods cannot consume events, for a fair comparison we reconstruct intensity frames from the events with E2VID \cite{rebecq2019events} and feed D-3DGS as additional input, which is reported as \emph{E2VID + D-3DGS}. We additionally compare against Event-boosted Deformable 3DGS (E-D3DGS) \cite{xu2025eventboosted}. To the best of our knowledge it is the only event-based deformable Gaussian method in this setting whose implementation is publicly available, as DEGS \cite{he2025degs}, Ev4DGS \cite{nakabayashi2025ev4dgs} and E-4DGS \cite{feng2025e4dgs} have released neither code nor data at the time of writing, so a direct comparison with them is not possible. Substituting the methods that cannot be reproduced by RGB baselines fed with E2VID reconstructions is common practice in this literature \cite{feng2025e4dgs}.

We further include an internal baseline, denoted \emph{Ours w/o Markov}, which disables the event-driven transition $g_\phi$ so that the events act only as supervision through the event loss. This variant follows the events-as-supervision paradigm of prior work and lets us isolate the gain brought by the event-driven Markov transition that is the core of EdMCGS.

\begin{figure}[t]
\centering
\includegraphics[width=\columnwidth]{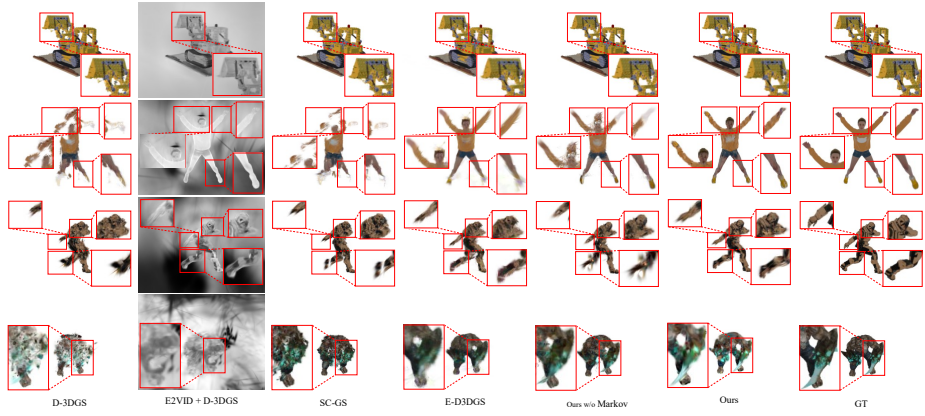}
\caption{Qualitative results on our synthesis dataset.}
\label{fig:qualitative_syn}
\end{figure}

\paragraph{Metrics}
We report the three standard image quality metrics, PSNR, SSIM, and LPIPS (VGG). E-D3DGS \cite{xu2025eventboosted} reports LPIPS with an AlexNet backbone by default, which we switch to VGG so that all methods are evaluated under the same metric.

\paragraph{Implementation details}
All experiments are conducted on a single GeForce RTX 4070 Ti Super with $16$\,GB of memory. Every method, ours and the baselines alike, is trained for $30$k iterations on the synthetic datasets and $40$k iterations on the real datasets. We use $M=512$ control points and bind each Gaussian to its $K=3$ nearest control points, while the temporal local isometry graph connects each control point to its $K_{\mathrm{conn}}=10$ neighbors. The event map is encoded by a shallow three-layer CNN into a $32$-channel feature map at a quarter of the input resolution, without any global pooling and with a receptive field of about $23$ pixels, so that each control point reads only its local neighborhood. The decoder $g_\phi$ is a small two-layer MLP whose last layer is zero-initialized; the exact layer specifications are given in our released code. The loss weights are $\lambda_d=0.2$, $\lambda_e=0.05$, $\lambda_t=10^{-5}$, and $\lambda_s=10^{-2}$.

\subsection{Comparison}
\label{subsec:comparison}

\begin{figure}
\centering
\includegraphics[width=\columnwidth]{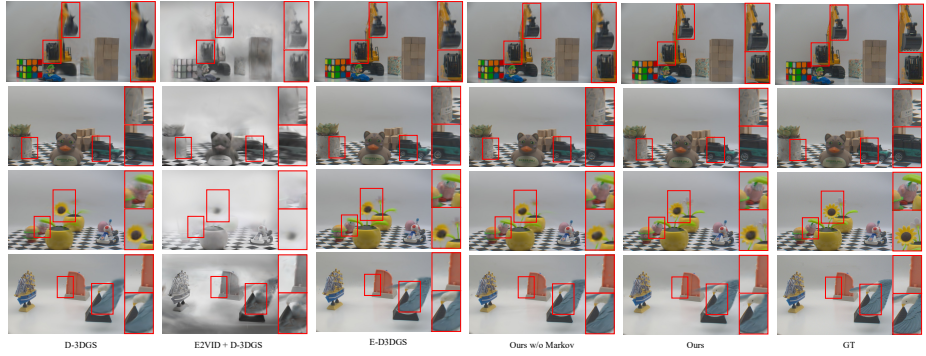}
\caption{Qualitative results on real-world dataset.}
\label{fig:qualitative_real}
\end{figure}

Figures~\ref{fig:qualitative_syn} and~\ref{fig:qualitative_real} show the same pattern in image space. The RGB-based methods either tear the moving parts apart or smear them into an artifact that overlays the positions of the two neighbouring observed frames, while E-D3DGS suppresses the tearing but leaves the fast-moving regions blurred, consistent with an approach whose stable evidence lies in the static background. Compared to previous methods, our method can reconstruct better details.

Quantitative results on our synthesis real-world datasets are presented in Table~\ref{tab:syn} and Table~\ref{tab:real}, the per-scene breakdown behind these averages is deferred to \ref{sec:appendix}. On our synthesis datasets, our method reaches $31.56$ dB at $10$ fps and $29.37$ dB at $5$ fps, exceeding all the baselines. Against E-D3DGS the relative gain is larger on LPIPS than on PSNR. This distinction matters here, because a method that cannot recover the motion inside an unobserved interval falls back on a temporally averaged, blurred rendering of the moving part, a failure that PSNR tolerates and LPIPS does not. On the four real scenes EdMCGS again ranks first on all three metrics. The absolute margins are smaller than on the synthetic data because these captures place a moderately sized moving object in an otherwise static scene, so much of every test image is explained by the static reconstruction and is never endangered by the low frame rate.

\begin{table}[H]
    \centering
    \caption{Quantitative comparison on our synthesis dataset, averaged over the four scenes. Per-scene results are given in \ref{sec:appendix} (Table~\ref{tab:syn_full}). We highlight the \textbf{best} and \underline{second-best} values for each column.}
    \label{tab:syn}
    \setlength{\tabcolsep}{4.5pt}
    \begin{tabular}{l ccc ccc}
    \toprule
     & \multicolumn{3}{c}{Average (5 fps)} & \multicolumn{3}{c}{Average (10 fps)}\\
    \cmidrule(lr){2-4}\cmidrule(lr){5-7}
    Method & PSNR$\uparrow$ & SSIM$\uparrow$ & LPIPS$\downarrow$ & PSNR$\uparrow$ & SSIM$\uparrow$ & LPIPS$\downarrow$\\
    \midrule
    D-3DGS \cite{yang2024deformable}   & 22.95             & .8975             & .1066             
                                       & 26.60             & .9302             & .0739\\
    E2VID + D-3DGS                     & 21.47             & .8923             & .1637 
                                       & 21.67             & .8925             & .1633\\
    SC-GS \cite{huang2024sc}           & 22.99             & .9052             & .1254             
                                       & 26.10             & .9204             & .0958\\
    E-D3DGS \cite{xu2025eventboosted}  & \underline{26.83} & \underline{.9298} & \underline{.0703}             
                                       & \underline{28.60} & \underline{.9546} & .0662\\
    Ours w/o Markov                    & 25.44             & .9280             & .0877
                                       & 28.11             & .9515             & \underline{.0624}\\
    \textbf{Ours}                      & \textbf{29.37}    & \textbf{.9495}    & \textbf{.0551}   
                                       & \textbf{31.56}    & \textbf{.9668}    & \textbf{.0426}\\
    \bottomrule
    \end{tabular}
\end{table}

The advantage grows as the input becomes sparser, which is the regime the method is built for. Halving the frame rate doubles the interval that must be filled without observations while leaving the event evidence inside it unchanged, so a method that draws its in-between motion from the events should degrade more slowly than one that interpolates over time. 

\emph{E2VID + D-3DGS} falls below our method in every setting, and its scores are almost invariant to the RGB frame rate, which shows that in extreme-low-frame-rate settings, the reconstructed frames dominate the optimization. Event-to-video reconstruction recovers relative brightness up to a drifting offset and carries no chrominance, so it is photometrically inconsistent with the genuine frames and corrupts the appearance those frames had already pinned down. 

E-D3DGS instead segments the scene into a dynamic and a static part and uses the static background as a photometric anchor, which works only while the moving object stays small in the image. The per-scene results in Table~\ref{tab:syn_full} follow that boundary, as our largest margin over E-D3DGS occurs at both frame rates on \emph{Jumpingjack}, the only sequence in which an articulated body fills the frame and leaves almost no static background to anchor on. EdMCGS needs no decomposition, no motion mask and no assumption about how much of the frame moves, and it ranks first on every scene at both frame rates.

\begin{table}
    \centering
    \caption{Quantitative comparison on the real-world dataset of E-D3DGS \cite{xu2025eventboosted} at 5 fps, averaged over the four scenes. Per-scene results are given in \ref{sec:appendix} (Table~\ref{tab:real_full}). We highlight the \textbf{best} and \underline{second-best} values for each column.}
    \label{tab:real}
    \begin{tabular}{l ccc}
    \toprule
    Method & PSNR$\uparrow$ & SSIM$\uparrow$ & LPIPS$\downarrow$\\
    \midrule
    D-3DGS \cite{yang2024deformable}  & 29.07             & .8829             & .2780\\
    E2VID + D-3DGS                    & 21.96             & .8165             & .5266\\
    E-D3DGS \cite{xu2025eventboosted} & 29.75             & .8920             & .2583\\
    Ours w/o Markov                   & \underline{30.22} & \underline{.8940} & \underline{.2575}\\
    \textbf{Ours}                     & \textbf{31.13}    & \textbf{.9015}    & \textbf{.2398}\\
    \bottomrule
    \end{tabular}
\end{table}

EdMCGS is at the same time the lightest of the event-based methods. As Tables~\ref{tab:efficiency_syn} and~\ref{tab:efficiency_real} show, it trains faster than E-D3DGS on both datasets, renders faster, and represents the scene with several times fewer Gaussians, while still reconstructing at higher quality. Relative to the D-3DGS backbone the event branch does cost training time and rendering speed, yet it leaves the number of Gaussians essentially unchanged, so the improvement comes from the motion model rather than from additional representational capacity. Event evidence is therefore better exploited by a light deformation model that reads it at inference than by a heavy scene representation that only absorbs it during training.

\begin{table}[t]
\centering
\caption{Efficiency comparison on our synthesis dataset.}
\label{tab:efficiency_syn}
\begin{tabular}{lcccc}
\toprule
Method & Time (mm:ss)$\downarrow$ & FPS$\uparrow$ & \#Gaussians & PSNR$\uparrow$\\
\midrule
D-3DGS \cite{yang2024deformable} & 5:49  & 566 & 7373  & 26.60 \\
E2VID + D-3DGS                   & 8:00  & 448 & 12097 & 21.67 \\
SC-GS  \cite{huang2024sc}        & 17:34 & 249 & 37699 & 26.10 \\
E-D3DGS \cite{xu2025eventboosted}& 15:04 & 277 & 46044 & 28.60 \\
\textbf{EdMCGS (Ours)}           & 13:34 & 309 & 8119  & 31.56 \\
\bottomrule
\end{tabular}
\end{table}

\begin{table}[t]
    \centering
    \caption{Efficiency comparison on real-world dataset (5 fps).}
    \label{tab:efficiency_real}
    \begin{tabular}{lcccc}
    \toprule
    Method & Time (mm:ss)$\downarrow$ & FPS$\uparrow$ & \#Gaussians & PSNR$\uparrow$\\
    \midrule
    D-3DGS \cite{yang2024deformable} & 14:02  & 296 & 18971  & 29.07 \\
    E2VID + D-3DGS                   & 18:38  & 221 & 23790  & 21.96 \\
    E-D3DGS \cite{xu2025eventboosted}& 198:58 & 176 & 183710 & 29.75 \\
    \textbf{EdMCGS (Ours)}           & 16:42  & 218 & 18051  & 31.13 \\
\bottomrule
\end{tabular}
\end{table}
\subsection{Ablation Study}
\label{subsec:ablation}

\begin{table}
    \centering
    \caption{Ablation of the main components of EdMCGS, with per-metric averages on the synthetic dataset (5 fps) and the real-world dataset. \textbf{Full} denotes the complete model.}
    \label{tab:ablation}
    \setlength{\tabcolsep}{4.5pt}
    \begin{tabular}{l ccc ccc}
    \toprule
     & \multicolumn{3}{c}{Synthetic (5 fps)} & \multicolumn{3}{c}{Real-world}\\
    \cmidrule(lr){2-4}\cmidrule(lr){5-7}
    Variant & PSNR$\uparrow$ & SSIM$\uparrow$ & LPIPS$\downarrow$ & PSNR$\uparrow$ & SSIM$\uparrow$ & LPIPS$\downarrow$\\
    \midrule
    w/o event stream & 24.05 & .8953 & .1015 & 30.04 & .8933 & .2533\\
    w/o Markov chain & 25.44 & .9280 & .0877 & 30.22 & .8940 & .2575\\
    w/o local sampling & 24.61 & .9033 & .0998 & 29.87 & .8899 & .2645\\
    w/o TLI          & 24.98 & .9220 & .0921 & 30.56 & .8985 & .2489\\
    \textbf{Full}    & \textbf{29.37} & \textbf{.9495} & \textbf{.0551} & \textbf{31.13} & \textbf{.9015} & \textbf{.2398}\\
    \bottomrule
    \end{tabular}
\end{table}

Table~\ref{tab:ablation} removes each of the three main components in turn. Discarding the event stream entirely, which reduces the model to a control-point deformation field driven by time alone, gives $24.05$ dB on the synthetic data and $30.04$ dB on the real data, $5.32$ dB and $1.09$ dB below the full model, and this difference measures everything the events are worth to our method.

The internal baseline \emph{Ours w/o Markov} keeps the control-point deformation field and the event loss but removes the event-driven transition, leaving the events as pure training-time supervision in the manner of prior work. It collapses to $21.67$ dB on \emph{Jumpingjack} at $5$ fps, $10.14$ dB behind the full model, yet trails by only $0.24$ dB on \emph{Lego} at $10$ fps. When little happens between two frames, conditioning the deformation on time alone is already close to sufficient, and it is when a great deal happens, which extreme-low-frame-rate capture guarantees, that the events must drive the transition at inference.

Replacing the local event sampling of Eq.~\eqref{eq:motioncode} by a global descriptor, obtained by average-pooling the encoded event map into a single vector shared by all control points, is worse still. The reason is that an accumulated event map mixes together the events fired by the moving object, by the static background under camera motion, and by sensor noise, and pooling collapses all of them into one descriptor that is then applied identically to every control point. Points on the static parts of the scene are pushed by the object's events and points on the object are dragged by the background's, which is precisely the confusion that reading the map at each point's own image projection avoids.

Removing the temporal local isometry term is comparably damaging. The event-driven transition produces an independent displacement for each control point, and the accumulated event map is noisy and unevenly distributed over the object, so nothing in the transition alone prevents neighbouring points from moving inconsistently. Preserving inter-point distances across nearby deformed timestamps is what keeps the propagated motion locally rigid, and the term matters most on the synthetic sequences, where the transition is doing the most work. The two components are therefore complementary rather than additive, since the transition supplies the motion and the isometry constraint makes it usable, and the full model needs both to reach its quality.

\section{Conclusion}
\label{sec:conclusion}

We present EdMCGS, a lightweight and end-to-end framework for reconstructing dynamic 3D scenes from extreme-low-frame-rate RGB images together with an event stream. Unlike previous methods that rely on temporal interpolation, the event-driven Markov chain remains active during inference, enabling intermediate deformations to be predicted directly from the event stream. Experiments on both synthetic and real-world datasets demonstrate that EdMCGS consistently outperforms existing baselines. We also release our code and dataset to facilitate future research on event-based dynamic scene reconstruction.

A limitation of EdMCGS is that our pipeline assumes known camera poses, an assumption that it shares with the deformable Gaussian splatting methods we compare against \cite{yang2024deformable, huang2024sc, xu2025eventboosted} and that DEGS \cite{he2025degs} lifts only by adding a separate pose network to its pipeline. Recent work reconstructs static scenes from events without poses \cite{huang2025inceventgs, kim2026e2egs}, and carrying that line over to a deforming scene, so that the camera poses and the dynamic geometry are recovered jointly, is a natural next step. Further directions include exploring higher-order event-driven temporal models and evaluating the method on larger real-world event datasets with more diverse motions.

\section*{CRediT authorship contribution statement}

\textbf{Yuzhong Wang}: Conceptualization, Methodology, Software, Validation, Investigation, Data curation, Writing -- original draft, Visualization. \textbf{Wenmin Wang}: Supervision, Writing -- review \& editing, Project administration. \textbf{Xinxing Yu}: Supervision, Writing -- review \& editing.

\section*{Declaration of Generative AI and AI-assisted technologies in the writing process}

During the preparation of this work the authors used Claude (Anthropic) in order to improve the language and readability of the manuscript. After using this tool, the authors reviewed and edited the content as needed and take full responsibility for the content of the published article.

\section*{Acknowledgements}

This research did not receive any specific grant from funding agencies in the public, commercial, or not-for-profit sectors.

\clearpage
\appendix
\section{Per-scene Results}
\label{sec:appendix}
\setcounter{table}{0}

For completeness, we report the per-scene breakdown behind the averages in Tables~\ref{tab:syn} and~\ref{tab:real}. Table~\ref{tab:syn_full} gives the per-scene results on the synthetic dataset, and Table~\ref{tab:real_full} gives the per-scene results on the real-world dataset. We highlight the \textbf{best} and \underline{second-best} values for each column.

\begin{table}[H]
    \centering
    \caption{Per-scene quantitative comparison on our synthesis dataset.}
    \label{tab:syn_full}
    \setlength{\tabcolsep}{4.5pt}
    \resizebox{\textwidth}{!}{%
    \begin{tabular}{l ccc ccc ccc ccc}
    \toprule
     & \multicolumn{3}{c}{Jumpingjack (5 fps)} & \multicolumn{3}{c}{Kick (5 fps)} & \multicolumn{3}{c}{Mutant (5 fps)} & \multicolumn{3}{c}{Lego (5 fps)}\\
    \cmidrule(lr){2-4}\cmidrule(lr){5-7}\cmidrule(lr){8-10}\cmidrule(lr){11-13}
    Method & PSNR$\uparrow$ & SSIM$\uparrow$ & LPIPS$\downarrow$ & PSNR$\uparrow$ & SSIM$\uparrow$ & LPIPS$\downarrow$ & PSNR$\uparrow$ & SSIM$\uparrow$ & LPIPS$\downarrow$ & PSNR$\uparrow$ & SSIM$\uparrow$ & LPIPS$\downarrow$\\
    \midrule
    D-3DGS \cite{yang2024deformable}   & 21.44             & .8850             & .1309             
                                       & 21.53             & .8955             & .1170
                                       & 25.86             & .8950             & .1152
                                       & 22.96             & .9146             & .0632      \\

    E2VID + D-3DGS                     & 19.12             & .8853             & .1717 
                                       & 22.14             & .9110             & .1681
                                       & 24.67             & .8975             & .1776
                                       & 19.94             & .8752             & .1374     \\

    SC-GS \cite{huang2024sc}           & 22.62             & .8944             & .1622             
                                       & 20.27             & .8919             & .1033
                                       & 25.19             & .8908             & .1752
                                       & 23.87             & .9437             & .0609\\

    E-D3DGS \cite{xu2025eventboosted}  & \underline{26.69} & \underline{.9161} & \underline{.0763}             
                                       & \underline{26.11} & .9276             & \underline{.0732}
                                       & 26.29             & .9190             & .0719
                                       & 28.23             & .9564             & .0599\\

    Ours w/o Markov                    & 21.67             & .9030             & .1253             
                                       & 25.14             & \underline{.9295} & .0826
                                       & \underline{26.49} & \underline{.9216} & \underline{.0916}
                                       & \underline{28.44} & \underline{.9580} & \underline{.0511}\\

    \textbf{Ours}                      & \textbf{31.81}    & \textbf{.9618}    & \textbf{.0422}    
                                       & \textbf{28.08}    & \textbf{.9423}    & \textbf{.0623}
                                       & \textbf{28.44}    & \textbf{.9332}    & \textbf{.0716}
                                       & \textbf{29.13}    & \textbf{.9608}    & \textbf{.0442}\\
    \midrule
     & \multicolumn{3}{c}{Jumpingjack (10 fps)} & \multicolumn{3}{c}{Kick (10 fps)} & \multicolumn{3}{c}{Mutant (10 fps)} & \multicolumn{3}{c}{Lego (10 fps)} \\
    \cmidrule(lr){2-4}\cmidrule(lr){5-7}\cmidrule(lr){8-10}\cmidrule(lr){11-13}
    Method & PSNR$\uparrow$ & SSIM$\uparrow$ & LPIPS$\downarrow$ & PSNR$\uparrow$ & SSIM$\uparrow$ & LPIPS$\downarrow$ & PSNR$\uparrow$ & SSIM$\uparrow$ & LPIPS$\downarrow$ & PSNR$\uparrow$ & SSIM$\uparrow$ & LPIPS$\downarrow$\\
    \midrule
    D-3DGS \cite{yang2024deformable}   & 24.83             & .9211             & .0801             
                                       & 24.51             & .9119             & .0982
                                       & 25.94             & .9116             & .0940
                                       & 31.11             & .9761             & .0232\\

    E2VID + D-3DGS                     & 19.12             & .8850             & .1715 
                                       & 22.14             & .9111             & .1683
                                       & 24.67             & .8974             & .1775
                                       & 20.76             & .8766             & .1358\\

    SC-GS \cite{huang2024sc}           & 24.65             & .9166             & .1343             
                                       & 23.59             & .8723             & .1349
                                       & 25.69             & .9187             & .0898
                                       & 30.46             & .9740             & .0241\\
                                       
    E-D3DGS \cite{xu2025eventboosted}  & \underline{28.42} & \underline{.9571} & \underline{.0738}             
                                       & \underline{28.21} & \underline{.9555} & \underline{.0619}
                                       & 27.45             & .9376             & .0820
                                       & 30.30             & .9681             & .0470\\

    Ours w/o Markov                    & 24.36             & .9443             & .0829             
                                       & 26.08             & .9363             & .0748
                                       & \underline{27.96} & \underline{.9429} & \underline{.0709}
                                       & \underline{34.03} & \underline{.9826} & \underline{.0208}\\

    \textbf{Ours}                      & \textbf{32.91}    & \textbf{.9707}    & \textbf{.0363}    
                                       & \textbf{29.70}    & \textbf{.9622}    & \textbf{.0538}
                                       & \textbf{29.35}    & \textbf{.9499}    & \textbf{.0628}
                                       & \textbf{34.27}    & \textbf{.9842}    & \textbf{.0175}\\
    \bottomrule
    \end{tabular}}
\end{table}

\begin{table}[H]
    \centering
    \caption{Per-scene quantitative comparison on the real-world dataset of E-D3DGS \cite{xu2025eventboosted} at 5 fps. The average over the four scenes is reported in Table~\ref{tab:real}.}
    \label{tab:real_full}
    \setlength{\tabcolsep}{4.5pt}
    \begin{tabular}{l ccc ccc}
    \toprule
     & \multicolumn{3}{c}{Excavator (5 fps)} & \multicolumn{3}{c}{Jeep (5 fps)}\\
    \cmidrule(lr){2-4}\cmidrule(lr){5-7}
    Method & PSNR$\uparrow$ & SSIM$\uparrow$ & LPIPS$\downarrow$ & PSNR$\uparrow$ & SSIM$\uparrow$ & LPIPS$\downarrow$\\
    \midrule
    D-3DGS \cite{yang2024deformable}  & 28.89             & .8953             & .3045             
                                      & 28.98             & .8665             & .2754\\
    E2VID + D-3DGS                    & 20.42             & .8239             & .5669             
                                      & 21.34             & .8085             & .5116\\
    E-D3DGS \cite{xu2025eventboosted} & \underline{30.88} & \underline{.9173} & \underline{.2528}    
                                      & 29.16             & .8747             & .2497\\
    Ours w/o Markov                   & 30.63             & .9107             & .2667 
                                      & \underline{30.03} & \underline{.8780} & \underline{.2490}\\
    \textbf{Ours}                     & \textbf{32.17}    & \textbf{.9270}    & \textbf{.2350}    
                                      & \textbf{30.49}    & \textbf{.8813}    & \textbf{.2402}\\
    \midrule
     & \multicolumn{3}{c}{Flowers (5 fps)} & \multicolumn{3}{c}{Eagle (5 fps)}\\
    \cmidrule(lr){2-4}\cmidrule(lr){5-7}
    Method & PSNR$\uparrow$ & SSIM$\uparrow$ & LPIPS$\downarrow$ & PSNR$\uparrow$ & SSIM$\uparrow$ & LPIPS$\downarrow$\\
    \midrule
    D-3DGS \cite{yang2024deformable}  & 27.40             & .8708             & .2681             
                                      & 31.01             & .8989             & .2639\\
    E2VID + D-3DGS                    & 20.95             & .8204             & .5145             
                                      & 25.15             & .8133             & .5134\\
    E-D3DGS \cite{xu2025eventboosted} & 27.85             & .8767             & .2673    
                                      & 31.12             & .8992             & \underline{.2633}\\
    Ours w/o Markov                   & \underline{28.57} & \underline{.8852} & \underline{.2431} 
                                      & \underline{31.66} & \underline{.9021} & .2711\\
    \textbf{Ours}                     & \textbf{29.51}    & \textbf{.8917}    & \textbf{.2237}    
                                      & \textbf{32.36}    & \textbf{.9060}    & \textbf{.2604}\\
    \bottomrule
    \end{tabular}
\end{table}

\bibliographystyle{elsarticle-num} 
\bibliography{ref}

\end{document}